\documentclass[runningheads]{llncs}
\usepackage{graphicx}
\usepackage[utf8]{inputenc} 

\usepackage{hyperref}
\usepackage{url}
\usepackage{float}

\usepackage[ruled]{algorithm2e} 

\usepackage[most]{tcolorbox}
\usepackage{caption}
\usepackage{subcaption}

\begin{document}
\title{Deep Active Learning in Remote Sensing for data efficient Change Detection}

\author{
  Vít Růžička, Stefano D'Aronco, Jan Dirk Wegner, Konrad Schindler \\
}\authorrunning{V. Růžička et al.}
%
\institute{EcoVision Lab - Photogrammetry and Remote Sensing, ETH Zürich \\
\texttt{previtus@gmail.com, \{firstname.lastname\}@geod.baug.ethz.ch}}

\maketitle

\begin{abstract}
We investigate active learning in the context of deep neural network models for change detection and map updating. Active learning is a natural choice for a number of remote sensing tasks, including the detection of local surface changes: changes are on the one hand rare and on the other hand their appearance is varied and diffuse, making it hard to collect a representative training set in advance.
In the active learning setting, one starts from a minimal set of training examples and progressively chooses informative samples that are annotated by a user and added to the training set. Hence, a core component of an active learning system is a mechanism to estimate model uncertainty, which is then used to pick uncertain, informative samples.
We study different mechanisms to capture and quantify this uncertainty when working with deep networks, based on the variance or entropy across explicit or implicit model ensembles.
We show that active learning successfully finds highly informative samples and automatically balances the training distribution, and reaches the same performance as a model supervised with a large, pre-annotated training set, with $\approx$99\% fewer annotated samples.

\end{abstract}

\section{Introduction}
\label{intro}

Mapping agencies as well as private companies world-wide are faced with the task of keeping their maps up-to-date over vast areas. While changes constantly occur and quickly add up to a large number -- buildings are erected or demolished, forest stands are cut down, etc.\ -- they concern only a tiny area of the overall territory. Maps are therefore not recreated from scratch, but renewed with incremental updates, mostly using data from periodic aerial or satellite surveys.
Arguably the biggest problem is to detect the changes that need to be introduced into the map:  adding or removing a few polygons to a map is not much work and can be done by a human cartographer. But to flag the sparse locations where a change has occurred one must, in principle, scan and inspect the entire area.
Consequently, a promising approach is to detect locations that have (potentially) changed with automatic image analysis and hand them to an expert for correction.

In recent years deep learning has emerged as a particularly powerful tool for image analysis, including remote sensing \cite{deep-remote-sensing} and change detection \cite{deep_chd_fcn_siam}, \cite{deep_chd_highRes}. 
Unfortunately, a well-known disadvantage of deep learning remains its hunger for large training sets. To supervise the training of high-capacity deep networks, many annotated examples are needed.
For the case of map updating (and several other remote sensing tasks), collecting that training data becomes a bottleneck. Changes are rare and sparsely distributed across the area of interest, moreover it is a-priori unclear which examples in what proportion are needed for a representative set.
That scenario suggests the use of active learning, where the model itself is used to automatically gather informative samples. In an iterative process, these samples are then labelled and added to the training set. By choosing the training examples with the highest chance to improve the preliminary model, one increases the sample efficiency and minimises the annotation effort.

In terms of machinery, active learning in its basic form is thus simply the repeated application of standard supervised learning. After every round, new data samples are selected and presented to an ``oracle" (in our case a human annotator) that adds the correct labels.
The one crucial component to make the learning system ``active" is an algorithm that can estimate how useful a (yet unlabelled) example will be to improve the model. Intuitively, these are examples where the model predictions are uncertain and therefore have a high chance of being wrong (since they are unlabelled it is not possible to decide with certainty).
The main challenge when using active learning in a deep learning framework boils down to estimating the model uncertainty of the predictions. While deep networks in principle output ``pseudo-probabilities", these scores are known to be badly calibrated, and additional measures are required.
Here, we focus on methods that are generic, in the sense that they can be easily added to most deep networks for computer vision and do not limit the user to a specific network architecture.
In particular, we investigate two different types of ensemble models: explicit ensembles of networks with the same architecture~\cite{power_of_ensembles_beluch2018power}, which differ due to the stochastic nature of the initialisation and optimisation processes used during training; and implicit ensembles that randomise certain parameters (such as dropout or batch normalisation) of the model at inference time, in this case we focus on Monte Carlo Batch Normalisation (MCBN)~\cite{MCBN}.

For our investigation, we use a real-world dataset of aerial images captured for the purpose (among others) of updating topographic maps.
The images cover the same geographic region in Switzerland but were collected several years apart. Co-registered patches are cropped from those images, and the task consists in labelling the pixels in a patch as changed or unchanged.
Overall there are 83'144 pairs. Like in most realistic scenarios the data are highly unbalanced, there are only 1'072 pairs with significant changes.
We show that with deep active learning only 1.02\% of them (850 pairs) must be annotated to achieve the same performance as a baseline model trained with knowledge of the full dataset (the baseline model was trained on a manually balanced subset, as otherwise the imbalance would degrade its performance; however, to select that subset, the labels of all available examples must be known).
Furthermore, we find that the choice of active learning method and uncertainty metric is less critical, as all models quickly reach the baseline performance. Still, explicit ensembles improve faster in early iterations and might be preferred when only few rounds of active learning are possible.

\section{Related Work}

\textbf{Active learning} is used as an umbrella term for frameworks where a learning system is trained in multiple steps and can choose which data points should be added to the training set in each step~\cite{cohn1996active}. 
The process that selects new samples from the dataset is often termed the \textit{acquisition function}, and it is generally assumed that an \emph{oracle} is available to annotate the new samples \cite{MCDROPOUT_4AL_gal2017deep}.
The survey \cite{settles2009active_learning_survey} gives an overview of traditional active learning methods based on expert knowledge or feature extractors such as SIFT.
Many active learning methods have been proposed in the context of remote sensing~\cite{AL_1}. In~\cite{Al_2} and~\cite{Al_3} the authors use a probabilistic framework based on Gaussian and Dirichlet processes to estimate the model uncertainty and select the data points to be labelled. In~\cite{Al_3} the authors tackle the problem of active learning in the context of remote sensing and domain shift. The proposed method uses a labelled dataset from the source domain and the optimal transport method to select which data points need to be labelled from a target domain for efficient learning.
Finally, in~\cite{Al_5} the authors conduct an extensive evaluation of several deep active learning methods on remote sensing datasets. They show the ability of Monte Carlo Dropout to effectively estimate model uncertainty in the context of active learning. The evaluation, however, does not focus on the scenario where the dataset is highly unbalanced, nor does it include \textit{Monte Carlo Batch Normalisation}.

\textbf{Uncertainty in deep networks. }
A key component of active learning is an effective representation of model uncertainty.
The work of \cite{ceal_wang2016cost} uses simple criteria derived from the model's prediction confidence to select samples. The disagreement between different networks in an ensemble has long been a popular measure of uncertainty \cite{NN_ensembles_Hansen1990}, \cite{ensemble_in_ML_dietterich2000ensemble}, and it has been shown to outperform other approaches for the task of image classification \cite{power_of_ensembles_beluch2018power}.
To avoid explicitly training multiple ensemble members, it has been suggested to use the dropout regulariser~ \cite{srivastava2014dropout} at inference time, so as to introduce stochasticity in the feed-forward prediction and obtain an implicit ensemble via \emph{Monte Carlo Dropout}~\cite{MCDROPOUT}, which later was also used for active learning~\cite{MCDROPOUT_4AL_gal2017deep}. %
A variant of the same idea is to instead randomise the batch normalisation layers~\cite{batch_norm_ioffe2015} in the network, leading to \textit{Monte Carlo Batch Normalisation} (MCBN)~\cite{MCBN}.
For our task, we aim to quantify uncertainty per pixel (rather than globally per image). This has so far been explored mostly in the context of semantic segmentation~\cite{al_yang2017suggestive,coal_sem_seg_gorriz2017cost,MCBN}. 

\section{Method}
\label{method}

Active learning aims to select samples from an unlabelled data set that, when labelled and added to the training set, bring the biggest benefit. The general sequence of operations in an active learning pipeline is shown in Alg.~\ref{active_learning_pseudocode}.


\begin{algorithm}[t] 
\SetAlgoLined
     \KwData{N\_iterations, InitialTrainingSet, UnlabeledData, N\_add}
     $TrainingSet = InitialTrainingSet$\;
     \For{$iteration\gets0$ \KwTo $N_{iterations}$}{
      train model on the $TrainingSet$\;
      run \textbf{acquisition function} on $UnlabeledData$ to select top $N_{add}$ samples\;
      $TrainingSet$ $\gets$ annotate and add selected $N_{add}$ data points\;
     }
     \caption{Pseudocode for active learning pipeline.}
      \label{active_learning_pseudocode}
\end{algorithm}

Arguably the most important part is the \emph{acquisition function}. It serves to estimate how informative a data point will be if it is labelled and added to the training set. With that function, one can simply rank the available data points and select the most informative ones.
Ideally, the acquisition function should estimate the expected performance improvement after adding a new point to the training set. Since that quantity is not tractable (as it involves retraining the model for every potential update) one must rely on some proxy that quantifies the model uncertainty at the 
new point. The intuition is that if the model is uncertain in some region of the input space, then adding a labelled example in that region will markedly reduce that uncertainty.

While the raw scores or logits of deep networks are notoriously over-confident and unreliable, there are multiple options for estimating model uncertainty. Here we focus on the two following methods:
\begin{description}
\item[Model Ensemble.] We train an ensemble of $M$ separate models with the same training set. Individual models differ due to randomness of the initial weights and to optimisation with stochastic gradient descent.
The disagreement between the $M$ predictions for the same, new data point is a well-established measure of uncertainty, albeit computationally somewhat inefficient.
\vspace{0.5em}
\item[Monte Carlo Batch Normalisation.] In this case only a single network is trained, but multiple predictions are obtained by randomising the batch normalisation (BN) inside the network. BN is normally a method to facilitate network training, by equalising the statistics over a batch of examples.
The MCBN method extends that idea to the prediction stage~\cite{MCBN}. With multiple rounds of inference, with randomly sampled normalisation parameters, it obtains a sort of ensemble prediction that can be seen as an approximation of the posterior.
Note that this only requires $M$ forward passes per point, possibly even re-using the early activation maps; whereas explicit ensembles train $M$ different complete models.
\end{description}
Given several sufficiently uncorrelated predictions for a sample, one can estimate the uncertainty of the model, and consequently the value of the acquisition function.  Here we test two different options as acquisition function: the variance metric \cite{al_yang2017suggestive} and the entropy metric~ \cite{MCDROPOUT,power_of_ensembles_beluch2018power}. 
In change detection the predictions are pixel-wise and as such we use the mean of these metrics over all pixels.
The variance metric is defined simply as the variance across the different predictions:  
\begin{equation}
\label{formula:variance_metric}
\sigma^2(x) = \frac{1}{MC} \sum_{c}\sum_{m}\big(p(y=c|x,w_m) - \hat{p}(y=c)\big)^2\;,
\end{equation}
where $x$ denotes the query data point, $w_m$ represents the parameters of the $m$-th model, and $\hat{p}(y=c)$ is the average prediction across all $M$ runs.
The entropy metric instead is defined as the entropy over the ``distribution" of predictions from different models:
\begin{equation}
\label{formula:entropy_metric}
 \begin{aligned}
    H[y|x, D_{train}] = - \sum_{c} \left( \frac{1}{M} \sum_{m} p(y=c|x,w_m) \right) 
    \cdot \log{ \left( \frac{1}{M} \sum_{m} p(y=c|x,w_m) \right)}\;.
 \end{aligned}
\end{equation}
Note, a large entropy can mean \emph{either} that all models are uncertain \emph{or} that they are certain, but in disagreement.
After computing the acquisition function, the unlabelled data are sorted by decreasing variance, respectively entropy, and a fixed-length list of the top $N_{add}$ points is passed to the oracle for labelling and added to the training set.\\

\begin{figure*}[tb]
	\centering
	\includegraphics[width=0.9\textwidth]{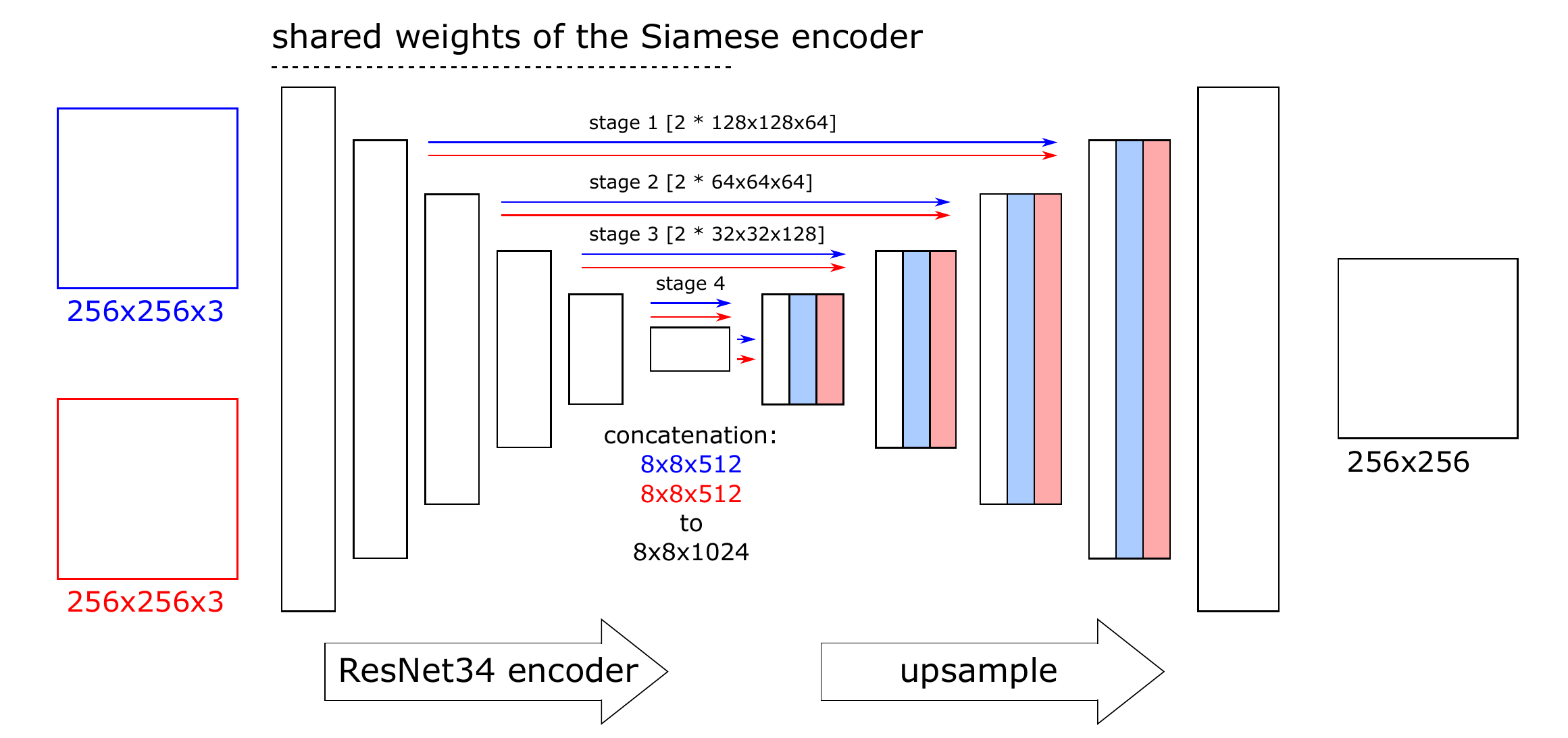}
	\caption{Siamese U-Net architecture with ResNet34 as encoder. 
	}
	\label{fig:model_architecture}
\end{figure*}

\begin{figure}[t]
	\centering
	\includegraphics[trim=0 0 0 0,clip,width=0.95\textwidth]{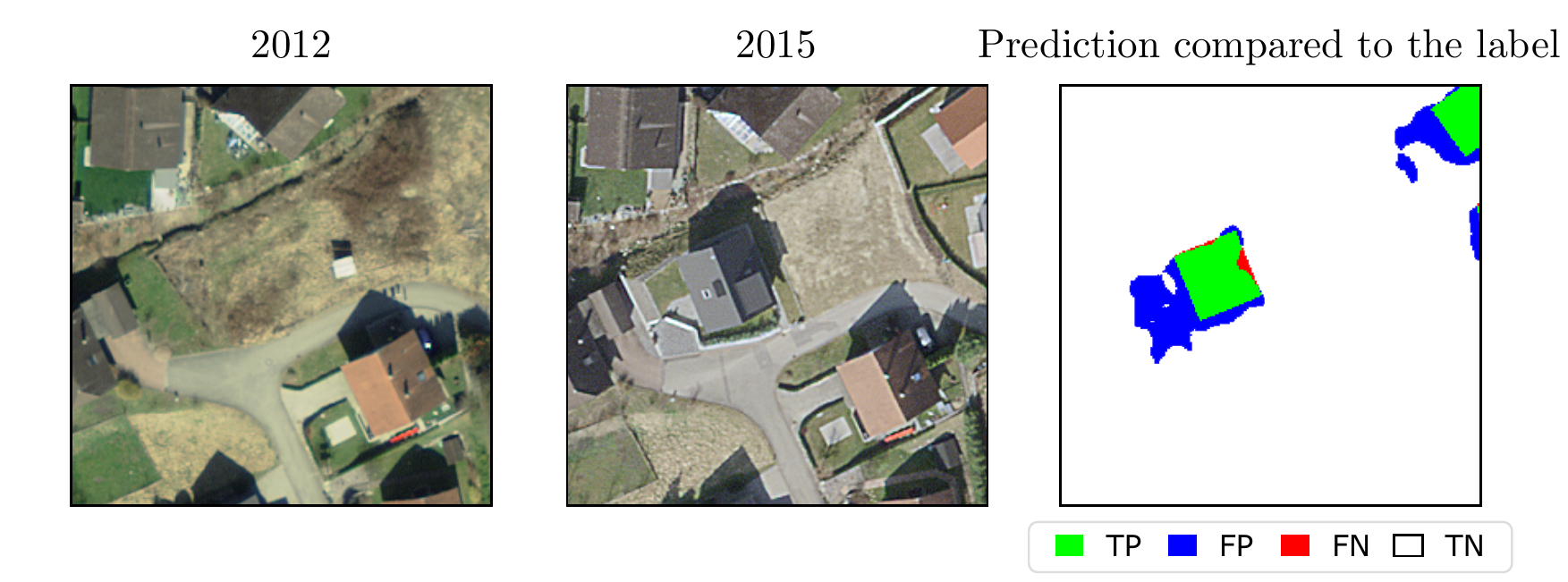}
	\caption{Example prediction of our Siamese U-Net, and comparison to ground truth. Following \cite{MCDROPOUT_4AL_gal2017deep,gutman2016skin_auc_sem_seg} we use the AUC metric, which averages across multiple thresholds, sidestepping the influence of any particular threshold setting.}
	\label{fig:prediction_example}
\end{figure}

\textbf{Network Architecture. }
Since our task is change detection between two images from different times, we use a Siamese network layout \cite{siamese_nns_similarity}.
I.e., the two inputs are encoded with two convolutional branches with all weights shared, the encodings are concatenated and then decoded back into a change map at the full input resolution.
We use a Siamese variant of U-Net, where the encoders are two identical copies of ResNet34 with shared encoder weights, see Figure \ref{fig:model_architecture}. 
The model returns a change map, where every pixel holds the probability that a change has occurred at the corresponding location. That map can be thresholded to obtain a discrete, binary change map. Figure \ref{fig:prediction_example} shows an example visualisation of the thresholded prediction and the possible types of errors.

Since ResNet34 is normally trained without dropout regularisation and we rely on pre-trained weights, we prefer not to employ \textit{Monte Carlo Dropout}~\cite{MCDROPOUT} and instead opt for MCBN.

\section{Experiments and Results}

\textbf{Dataset. }
We work with a collection of aerial images with a ground sampling distance of 25 cm, captured over an area of 303 km$^2$ in the Aargau canton of Switzerland. Two sets of images were acquired in 2012 and 2015 and aligned by ortho-rectification. The data was provided by the Swiss national mapping agency (Swisstopo).
We tiled the area of interest into a total of 83'144 tiles of size 256$\times$256 px, with 32 px overlap to mitigate boundary effects.
For that area we also obtained pixel-accurate annotations with two classes ``changed" or ``unchanged" (only changes to buildings are annotated).

\begin{figure}[t]
	\centering
	\includegraphics[trim=0 25 0 0,clip,width=1.0\textwidth]{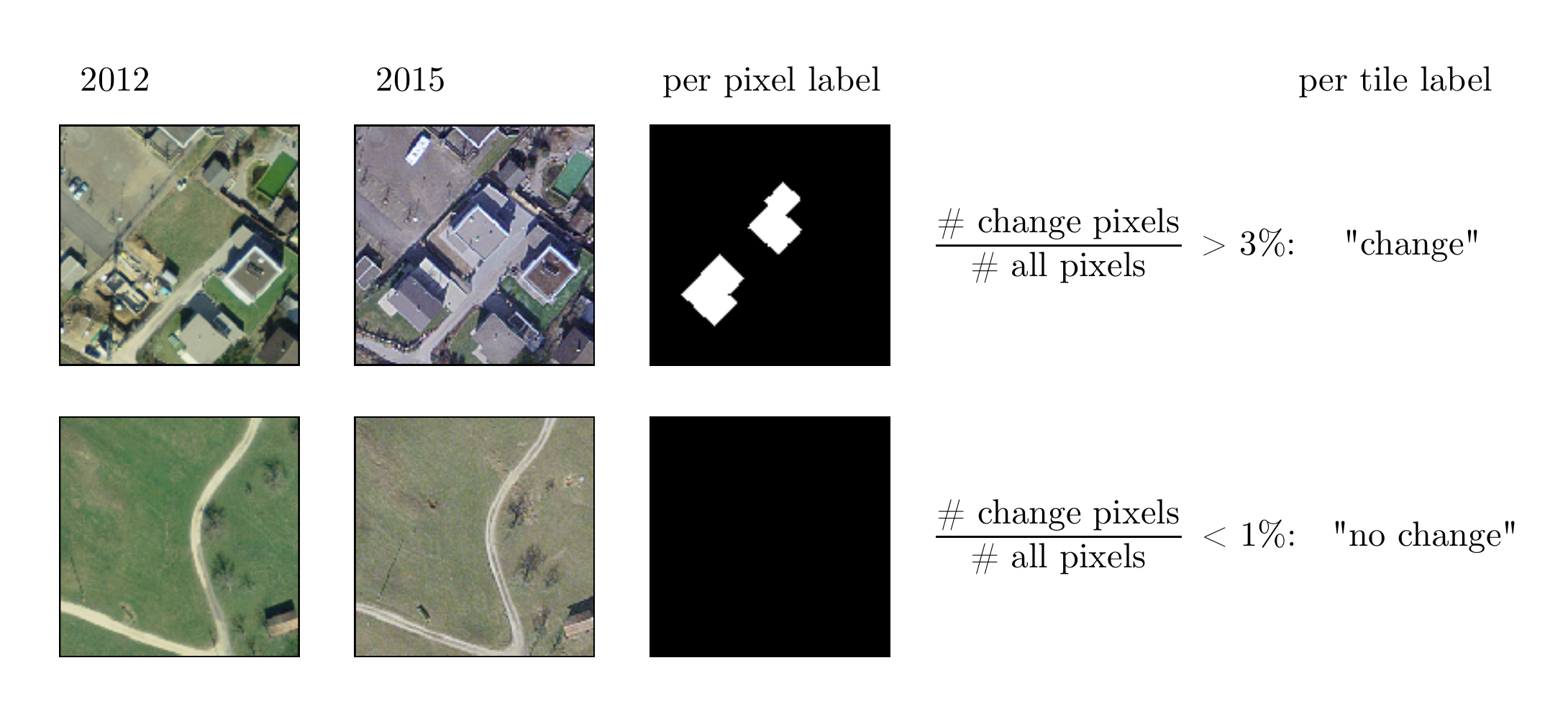}
	\caption{Example pairs of images tiles for change detection, and associated ground truth label maps.
	}
	\label{fig:data_example}
\end{figure}

To better assess the overall number of change events we have calculated per-tile labels, where tiles with \textgreater3\% changed pixels are considered as changed and tiles with \textless1\% changed pixels are considered unchanged. Values between 1 and 3\% are ignored to account for boundary effects and label noise. See Figure \ref{fig:data_example} for an example.
As expected the dataset is highly unbalanced, there are only 1'072 tiles with changes. To speed up training we discard 40'000 unchanged examples, such that the final dataset contains 42'072 unchanged tiles. We point out that the experimental findings are nevertheless equally valid for the full dataset of 83'144 tiles, as the bottleneck is to localise and annotate all 1'072 changes.

Note that in a fully supervised standard setting the training set would usually be balanced, by sub-sampling the unchanged tiles and/or synthetically augmenting the changed ones.
These strategies are not applicable in the active learning scenario, where only a minimal initial set of labels is available (for both classes).\\

\textbf{Experimental setup and evaluation metrics. }
For all our experiments we use the Siamese U-net, with ResNet34 encoder pre-trained on ImageNet. We fine-tune the model for 100 epochs using the Adam optimiser, base learning rate of $10^{-5}$ and weighted cross-entropy loss, with 3$\times$ higher weight for the ``change" class. Moreover, we employ data augmentation with the following transformations: flip horizontally, flip vertically, rotation by 90$^\circ$, rotation by 270$^\circ$.

To emulate an active learning scenario we always start with a balanced initial set of 25 changed and 25 unchanged tiles.
The remaining 43'094 tiles form the ``unlabelled" dataset from which the active learning loop selects additional samples in batches of $N_{add}=100$.
Model performance is measured using the area-under-the-curve (AUC) metric computed over individual per-pixel labels on a set-aside, balanced test set of 200 images, as in \cite{MCDROPOUT_4AL_gal2017deep,gutman2016skin_auc_sem_seg}. To monitor the sample selection we also record the class distribution every time the training set size is increased.

As a baseline and ``upper bound" we also train the same network architecture outside of the active learning loop, using a balanced training set of all 1072 changed tiles and 1072 randomly sampled unchanged tiles.
Note, the balancing is done only to avoid that the large number of unchanged samples overwhelm the changed ones and degrade the model.
It does not reduce the annotation effort, which consists entirely in finding and segmenting the changed tiles. Once all changes have been annotated, unchanged examples are trivial to find by randomly sampling the remaining area.

We also include a naive active learning baseline as ``lower bound" and sanity check, where model uncertainty is not estimated, instead the acquisition function is purely random. At each iteration of the active learning loop, we randomly sample 100 still unused tiles into the training set.
Note that in this case, the training set converges towards the unbalanced distribution of the overall data.
Since no uncertainty must be calculated, one only needs to train a single model without MCBN for this baseline. However, while computationally more efficient, this would sacrifice the benefit of ensemble averaging. We therefore run both versions.\\

\textbf{Results. }
We test four different active learning variants (explicit ensembling and MCBN, each with variance or entropy metric as acquisition function) and compare them to the two baselines described above.

\begin{figure}[t]
	\centering
\begin{subfigure}[t]{0.48\linewidth}
    \centering
    \includegraphics[width=\textwidth]{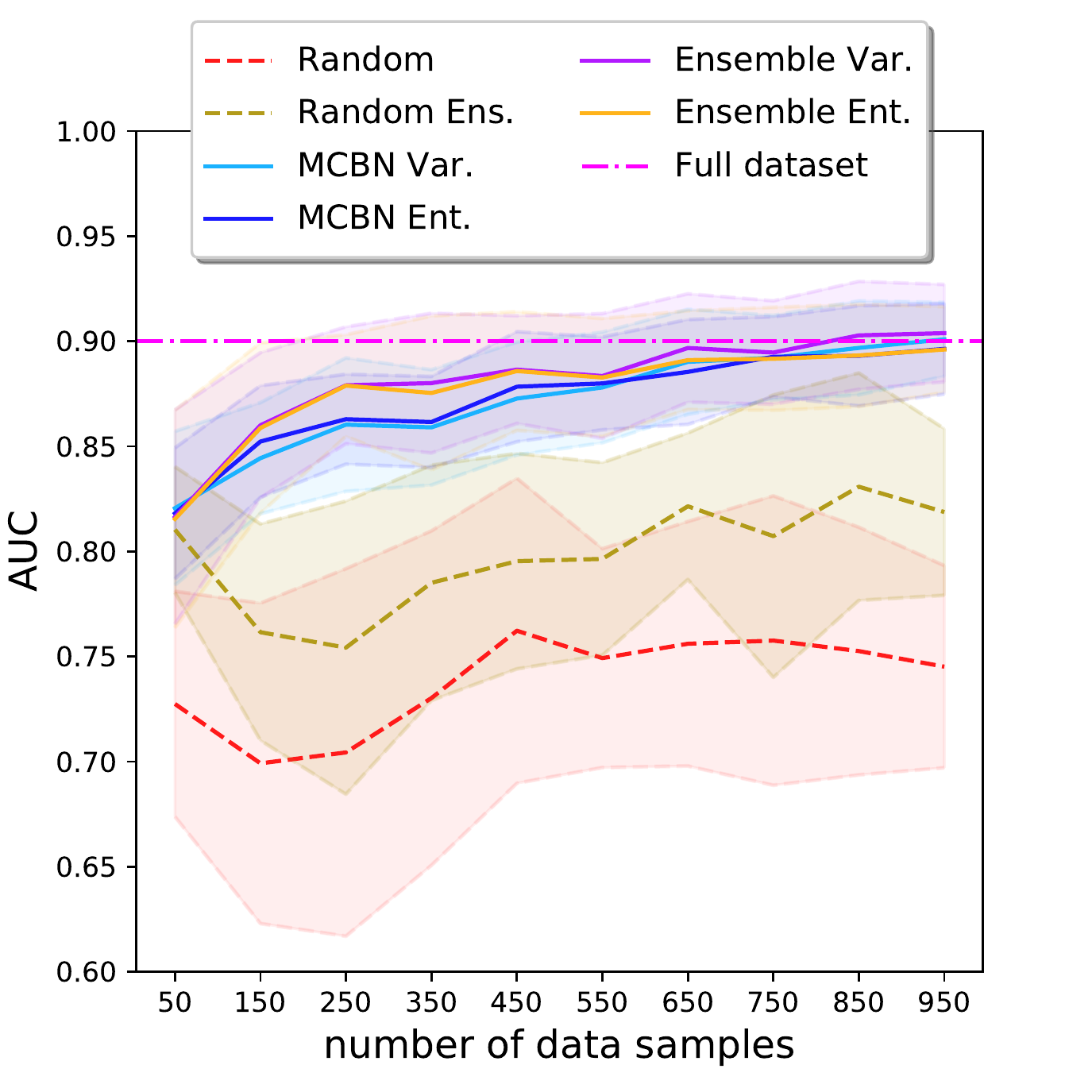}

\end{subfigure}
\hfill
\begin{subfigure}[t]{0.48\linewidth}
    \centering
    \includegraphics[width=\textwidth]{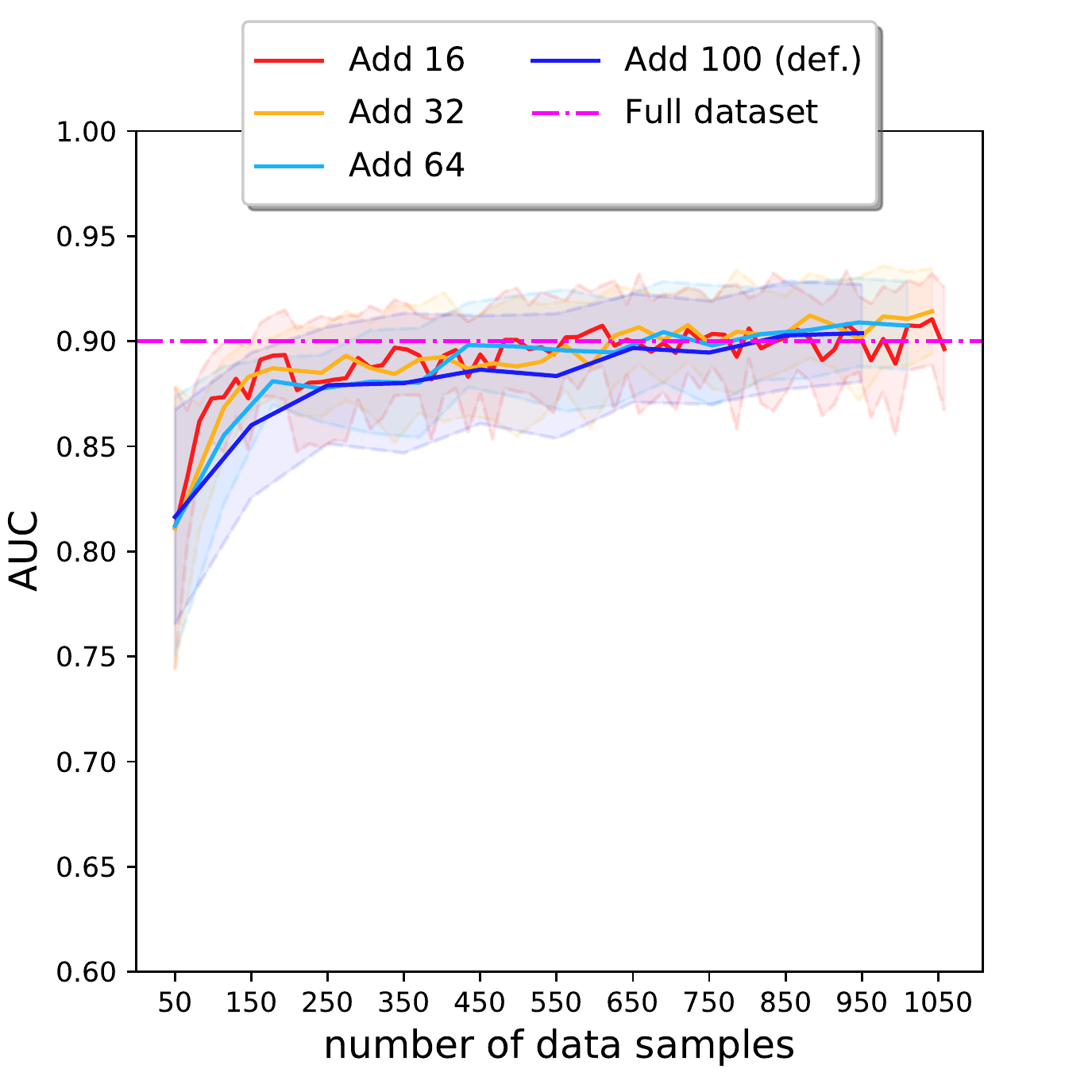}

\end{subfigure}
	\caption{\label{fig:auc_plot}
	(\emph{left}) Comparison of different sample selection metrics. For each metric, we iteratively add the top 100 selected training samples per round and plot the mean AUC and its standard error over 20 runs.
	(\emph{right}) Ablation study for top-$k$ samples per iteration. The experiment was run with an ensemble of 5 models and the variance metric. We plot mean AUC and standard error over 10 runs.
	}
	
\end{figure}

\begin{figure}[t]
	\centering
\begin{subfigure}[t]{0.32\linewidth}
    \centering
    \includegraphics[width=\textwidth]{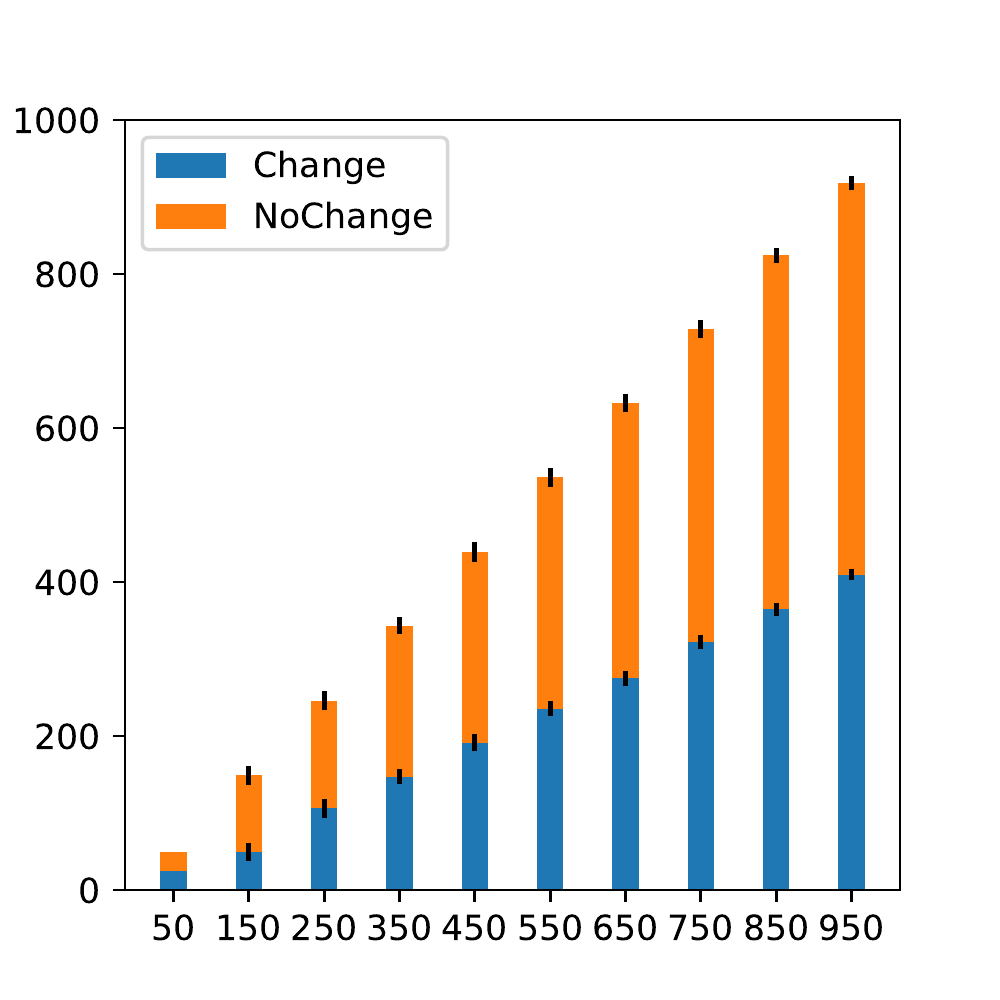}
    \caption{Ensemble Variance}

\end{subfigure}\hfill
\begin{subfigure}[t]{0.32\linewidth}
    \centering
    \includegraphics[width=\textwidth]{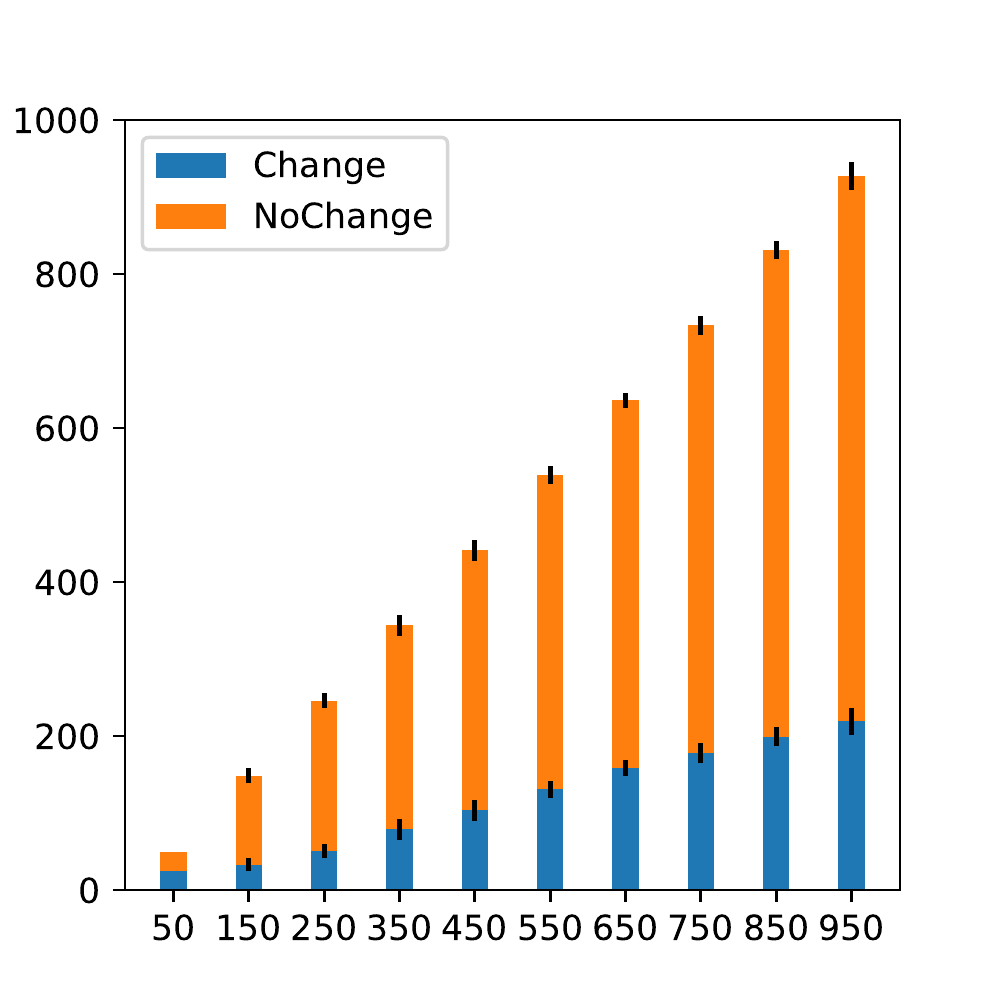}
    \caption{MCBN Variance}

\end{subfigure}\hfill
\begin{subfigure}[t]{0.32\linewidth}
    \centering
    \includegraphics[width=\textwidth]{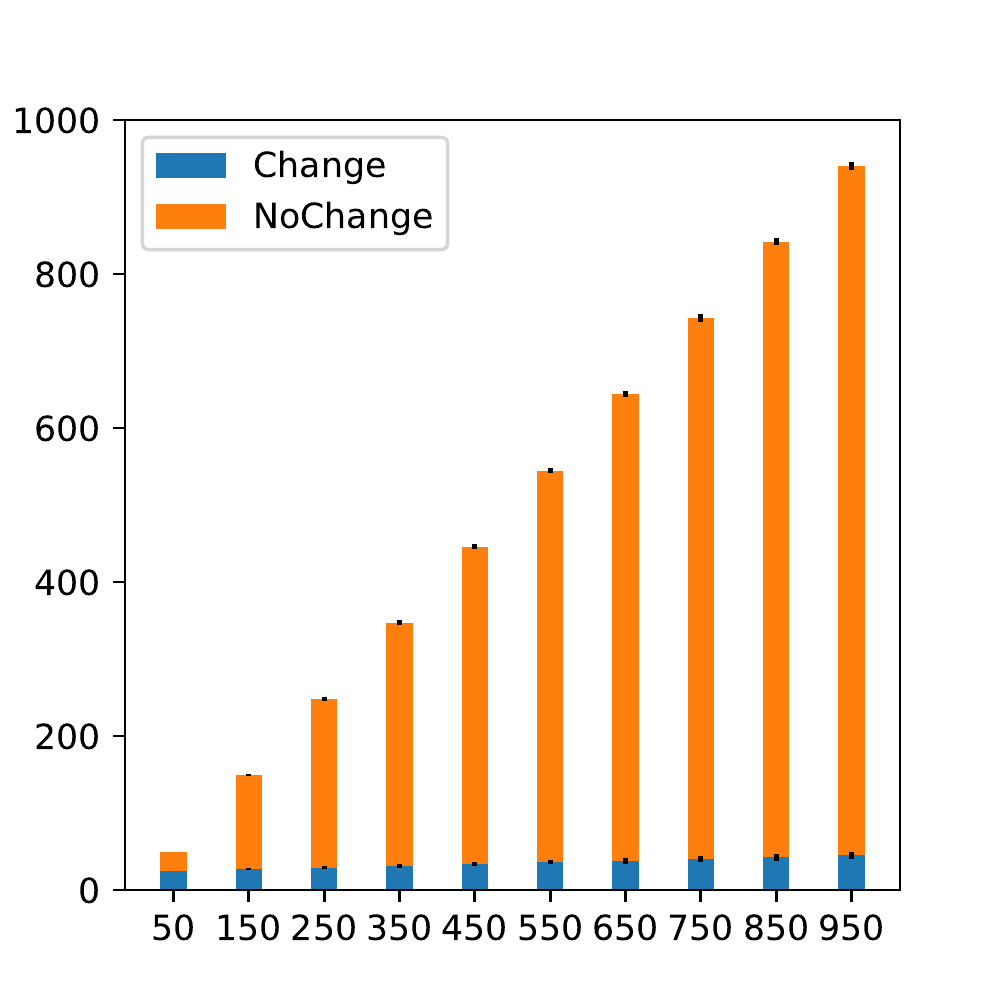}
    \caption{Random}

\end{subfigure}

	\caption{Label distribution over active learning iterations (averaged over 20 runs for each experiment). The acquisition function implicitly ``mines" for samples from the rare ``change" class.} \label{fig:balances}
	
\end{figure}

In Fig.~\ref{fig:auc_plot}(left) we show a comparison between all tested methods. Ensembles and MCBN models were always trained with $M=5$ networks, respectively stochastic forward passes.
Every experiment was repeated 20$\times$. The lines in the graph denote the mean AUC values over 20 runs, the shaded area indicates the standard deviation.
Any reasonable active learning method clearly outperforms the random baseline.
In general, the tested methods behave similarly and at a training set size of $\approx$850 tiles match the upper bound supervised with all tiles. While they reach comparable performance, the explicit model ensemble outperforms MCBN in the early rounds of active learning.
The difference between the variance and entropy metrics is negligible, if anything the theoretically less powerful variance metric
seems to have a tiny (but not statistically significant) edge.
The random baseline performs very poorly, mostly due to the fact that in the unbalanced dataset it hardly picks up any examples from the ``change" class. On the contrary, the more qualified acquisition functions rank the rare ``change" samples as more informative on average and inject a large portion of them into the training set. I.e., active learning can also be seen as a measure to ensure that a small training set is sufficiently balanced. In Figure \ref{fig:balances} that effect is illustrated for the variance metric (graphs for entropy look similar). The effect is more pronounced with the explicit ensemble, which ends up with more than 40\% samples with changes. With MCBN only about 20-22\% of the sampled tiles contain changes, likely because the $5$ predictions are more correlated. Still, that proportion is enough to attain the upper bound supervised with all 1072 changed samples.
Note also the error bars, which show that the selected class proportions are extremely stable across runs.

We have also conducted an experiment where we change the amount of training samples added per iteration of the active learning loop (using the variance metric). See Fig.~\ref{fig:auc_plot}(right). In order to provide comparable results, we also adjust the number of iterations, such that every experiment ends up with at least 950 training samples.
As a general trend, many iterations with small samples increase the performance faster in the early stages, but at the price of a noisier and less monotonous increase.
In this context, it should be mentioned that the choice of $N_{add}$ is also subject to practical considerations, as small steps require many more repetitions of model training, and may be inconvenient due to the many annotation sessions, each with very few samples.

\section{Conclusion}
We have investigated active learning with contemporary deep network models for change detection in remote sensing data.
We have shown that, by actively selecting informative samples, one can reach the same detection performance with a fraction of the labelling effort.

In our experiments, different methods to estimate the prediction uncertainty and different acquisition functions have performed comparably well. All qualified methods clearly outperform a naive random baseline and, with only a few hundred samples, reach the performance of a model trained on the full available dataset.
Importantly, our experiments have confirmed that active sample selection tends to automatically balance the training set, despite an extreme class imbalance in the input data.
Our experiments also support the claim that stochasticity at test time -- like in our case MCBN -- is a valid, more efficient proxy for the well-proven explicit ensemble model. We did however observe some differences in their behaviour, a point that may deserve further study.

\bibliographystyle{splncs04} 
\bibliography{biblio}

\end{document}